\documentclass{article}

\usepackage{PRIMEarxiv}

\usepackage{url}
\usepackage{xcolor}
\definecolor{newcolor}{rgb}{.8,.349,.1}

\usepackage{tabularx}
\usepackage{standalone}
\usepackage{graphicx}
\usepackage{booktabs}
\usepackage{gensymb}
\usepackage{hyperref}
\usepackage{longtable}
\usepackage[normalem]{ulem}
\usepackage{float}
\usepackage{svg}
\usepackage{breakurl}
\usepackage{refcount}
\usepackage{url}
\usepackage[flushleft]{threeparttable}
\DeclareMathAlphabet{\pazocal}{OMS}{zplm}{m}{n}

\usepackage[utf8]{inputenc} 
\usepackage{amsmath,amsthm,amsfonts,dsfont,mathtools} 
\usepackage{longtable}

\title{Visual Question Answering: A Survey on Techniques and Common Trends in Recent Literature
}

\author{
  Ana Cláudia Akemi Matsuki de Faria*\\
  Eldorado's Institute of Technology, Brazil \\
  \texttt{ana.faria@eldorado.org.br}\\
  \AND Felype de Castro Bastos*\\
  Eldorado's Institute of Technology, Brazil \\
  \texttt{felype.bastos@eldorado.org.br}\\
  \AND José Victor Nogueira Alves da Silva*\\
  Eldorado's Institute of Technology, Brazil \\
  \texttt{jose.silva@eldorado.org.br}\\
  \AND Vitor Lopes Fabris*\\
  Eldorado's Institute of Technology, Brazil \\
  \texttt{vitor.fabris@eldorado.org.br}\\
  \AND Valeska de Sousa Uchoa\\
  Eldorado's Institute of Technology, Brazil \\
  \texttt{valeska.uchoa@eldorado.org.br}\\
  \AND Décio Gonçalves de Aguiar Neto\\
  Eldorado's Institute of Technology, Brazil \\
  \texttt{decio.neto@eldorado.org.br}\\
  \AND Claudio Filipi Goncalves dos Santos\\
  Eldorado's Institute of Technology, Brazil \\
  \texttt{claudio.santos@eldorado.org.br} \\
}

\begin{document}
\maketitle
\begin{abstract}
\it Visual Question Answering (VQA) is an emerging area of interest for researches, being a recent problem in natural language processing and image prediction. In this area, an algorithm needs to answer questions about certain images. As of the writing of this survey, 25 recent studies were analyzed. Besides, 6 datasets were analyzed and provided their link to download. In this work, several recent pieces of research in this area were investigated and a deeper analysis and comparison among them were provided, including results, the state-of-the-art, common errors, and possible points of improvement for future researchers.
\end{abstract}


\footnote{*Authors contibuted equally to this research.}

	\section{Introduction}
	\label{s.introduction}
{\it Visual Question Answering}	 (VQA) is a multi-disciplinary artificial intelligence research problem that has attracted the attention of researchers from computer vision, natural language processing, knowledge representation, and other machine learning communities. To solve that question, VQA is a task of generating natural language answers when a question in natural language is asked related to an image.
In recent years, visual question answering as a result of the flourish in this field, datasets, metrics, and models have been proposed, and the scope of research has been expanded.
Although artificial intelligence has solved several different problems, such as image classification and natural language processing (NLP), it is hard to model a problem which needs different types of data. For instance, mixing computer vision with NLP to retrieve some information about an image from a question has tricked researchers for several years.

\begin{figure}[b]
	\centering
	\includegraphics[scale=0.35]{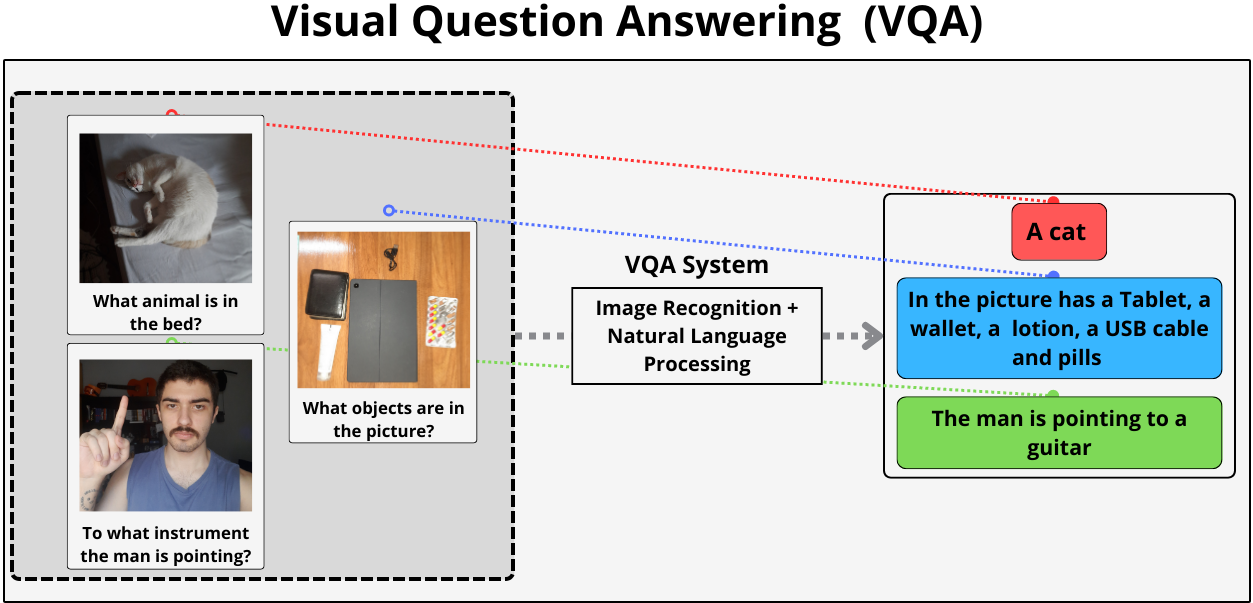}
	\label{fig.vqa}
	\caption{Example of how a VQA system works}
\end{figure}
	
	\subsection{Comparison with other works}
	
Few surveys have been developed about VQA, the different approaches to solving this task and developing new data to complement existing benchmarks. The majority of them focuses on providing a taxonomic structure for models and datasets applied, with some regarding specific sub-areas of VQA~\cite{lin2022medical} while others tend to a more general description of the subject~\cite{sharma2021survey, manmadhan2020visual,patil2020visual,Zou2020}. A general comparison is provided in this section, considering the present work and other surveys in the field, following the research premises established in section \ref{ss.scope_of_this_work}.

The work conducted in \cite{sharma2021survey} brings a general overview of this research area, discussing classical, well-established VQA datasets alongside new ones, evaluation metrics to understand and measure different aspects of a model, and multiple architectures used to solve VQA (including those where extra information is provided, such as scene-text, present in specific datasets \cite{TextVQA, STVQA_original}).

\cite{manmadhan2020visual} reviews the state-of-the-art in VQA whilst providing step-by-step descriptions and explanations as to current methodologies, datasets and evaluations. A critial analysis of the state of the field is discussed, as well as possible future directions.

In \cite{patil2020visual}, the authors discuss the state-of-the-art in Visual Question Generation (VQG), an important complementary task of VQA, specially concerning the development of novel datasets. They review the current methodology involved in VQG, how to measure the adequacy of generated questions, common algorithms in the subfield, as well as provide a discussion for existing challenges.

\cite{lin2022medical} discusses a comprehensive overview of the techniques and approaches used to answer questions related to medical images. This survey also explores specific datasets in the health field and analyzes approximately 45 papers.

\cite{Zou2020} reviews the existing datasets, metrics, and models of VQA and analyzes their progress and remaining problems.

This survey discussed the steps involving VQA task with recent papers and evaluate them with their different approaches to solve the same problem.
	
	\subsection{Scope of this work}
	\label{ss.scope_of_this_work}
The above-mentioned surveys have contributed to a better analysis of different approaches to solving the VQA task, exploring the medical area applied to VQA, and developing new data to complement existing benchmarks. The majority of them focus on providing a taxonomic structure for models and datasets applied, while others tend to a more general description of the subject. 

This paper presents a comprehensive exploration of the VQA task, analyzing recent papers and evaluating their different approaches to address the same problem. Additionally, we delve into crucial aspects, such as bias reduction, integration of external knowledge, and the utilization of transformers for both visual and textual enhancements. The major contributions of this survey are:
	
\begin{itemize}
	\item Novelty: to introduce the most recent and significant works comprising strategies for replacing parts or the entire VQA pipeline through deep learning approaches;
	\item Recently developed: all studies considered were published between the years 2020
	and 2022, making this study very up-to-date;
	\item New tendencies: in this study, many papers follows tendencies that are described in \ref{ss.tendencies};
	\item Finally, all works analyzed are open-access, facilitating their reproducibility for the scientific community.
\end{itemize}
{The papers gathered and reviewed in the present survey are described in Table \ref{tab.descriptionworks}.
		
		\begin{longtable}{|c|p{0.8\linewidth}|}
			\caption{Description of the papers selected for this survey}
			\label{tab.descriptionworks} \\
	\hline
			Reference & Description \\ \hline
			\endfirsthead
			\endhead
			\cite{yang2020learning} & Develops two architectural branches to deal with bias learning: Content, negatively affected by bias, and Context, positively affected by bias \\ \hline
			
			\cite{vath2021beyond} & Novel digital framework to analyze, evaluate and test models and datasets \\ \hline
			
			\cite{farinhas2021multimodal} & Models attention mechanisms of VQA models as multimodal feature functions for closer resemblance to human attention \\ \hline
			
			\cite{kervadec2021roses} & Proposes a novel dataset and evaluation procedure for out-of-distribution benchmarking of models' generalization capacities \\ \hline
			
			\cite{teney2020unshuffling} & Implements data partitions and training environments to reduce spurious correlations while maintaining concrete ones \\ \hline
			
			\cite{gao2021structured} & Novel modules for reading text in images on VQA and new annotations for the TextVQA dataset \\ \hline
			
			\cite{chen2021zero} & Zero-shot modeling of VQA algorithms using external knowledge graphs with a new dataset for this task  \\ \hline
			
			\cite{zhan2020medvqa} & Proposes a question-conditioned reasoning module for medical VQA models \\ \hline
			
			\cite{vatsal2020iqvqa} & Implements a model-agnostic implication generator for models to deal with logical inferences when answering \\ \hline
			
			\cite{spencer2021skill} & Develops a new way to evaluate VQA models following "skills and concepts" present in the image  \\ \hline
			
			\cite{wen2021debiased} & Proposes a technique to explicitly recognize and alleviate the effects of negative bias during training \\ \hline
			
			\cite{rajar2021graphhopper} & Proposes a decomposition of visual concepts in a graph structure for more contextualized answers \\ \hline
			
			\cite{claudio2019psycholinguistics} & A psycholinguistic approach to understanding and dealing with VQA models catastrophic forgetting \\ \hline
			
			\cite{wang2021mirtt} & Development of a trilinear model to account not only for images and questions, but also information contained in the answers \\ \hline
			
			\cite{liu2021constrative} & Contrastive learning and posterior representation distillation for unsupervised learning of radiology images in a VQA setting \\ \hline
			
			\cite{kafle2020answering} & Implements a simple, yet efficient, bimodal fusion technique for the task of CQA \\ \hline
			
			\cite{nguyen2022coarse} & Proposes an embedding framework for region-of-interest and predicate extraction from image-question pairs \\ \hline
			
			\cite{hu2021question} & Proposes a novel task consisting of scene text-aware automatic image caption generation, along with two datasets and a model to solve them \\ \hline
			
			\cite{gong2021sysu} & Documents their participation in the VQA-Med competition of 2021, as well as the model developed  \\ \hline
			
			\cite{Liu2022} & Proposes a regularization in attention layers, improving the visual information extraction \\ \hline
			
			\cite{Salewski2022} & Proposes an enrichment of the CLEVR dataset with more detailed synthetically-generated annotations \\ \hline
			
			\cite{https://doi.org/10.48550/arxiv.2111.02358} & Proposes a way to train models for VQA in a separate way, combining individually trained models in a way to achieve this \\ \hline
			
			\cite{Han2021} & Reduction of model's bias learning by overfitting biased data and fine-tuning on unbiased data \\ \hline
			
			\cite{Liang2021} & Develops an objective function based on the question's language-prior to balance the loss outcome of biased functions \\ \hline
			
			\cite{Qu2021} & Proposes a dual-encoder dense retrieve to imbue VQA models with external unstructured information \\ \hline
			
            \cite{LaTr} & Using Scanned Documents, authors uses transformers to solve ST-VQA \\ \hline
	
			\cite{MUSTVQA} & A transformer-based model capable of answer question from different languages\\ \hline
			
			\cite{gong2021sysu} & An interactive visual analytics tool for vision and language reasoning in transformers.\\ \hline	
			
			\cite{kervadec2021transferable} & Increase in robustness of VQA models through transfer learning of a perfect-sighted model \\ \hline
			
			\cite{Khan2022WeaklySG} & Visual feature extraction improvement for increased VQA performance using transformers \\ \hline
			
	\end{longtable}
	}
	
		\subsection{Papers analysis}
	\label{ss.papersanalysis}
	
	{The methodology to analyze each work reviewed in this survey is described here, focusing on highlighting points of improvement, methodologies used, and strategies for further development in the field of Visual Question Answering (VQA). The key points used for the analysis of the works presented in this survey are:}
	{\begin{itemize}
			\item \textbf{Introduction}: During the introduction, it is possible to get insights about the proposed paper and make a general analysis of the topic;
			\item \textbf{Methodology}: Observing the methodology of the papers, it is possible to obtain the problems that were proposed as well as their resolution;
			\item \textbf{Datasets}: Datasets are important inclusion factors; their evaluation demonstrates the model's capability;
			\item \textbf{Results achieved}: When the author analyzes his results, he brings an opinion based on reported facts that can help in the understanding and impact of his work;
	\end{itemize}}

	\subsection{Survey Overview}
	
	{The remainder of this survey is organized as follows: Section \ref{s.deep_explanation} gives deep explanations about VQA, how it works, applications, tendencies, and most common methods. Section \ref{s.known_approaches} reviews all articles analyzed in this survey, following a structure to contemplate highlights of the most recent papers in last years in VQA. Section \ref{s.methodology} describes the datasets and metrics used and to complete this comparative analysis, Section \ref{s.results} gives a comparison results to evaluate them to bring conclusions. Section \ref{s.discussion} discusses the work and contemplates the state-of-the-art, common points in the articles and an exploration in the weakness in these works. Section \ref{s.conclusion} concludes this study with evaluations, giving directions for future work and final considerations.}

\section{Visual Question Answering (VQA)}
\label{s.deep_explanation}

The VQA task consists of accurately answering an image-question pair $(I, q)$ based on characteristics of both $I$ and $q$. Although formulations of this task have appeared prior to 2015 as a collision between image recognition, natural language processing and knowledge representation, it gained notoriety with the publication of \cite{VQAoriginal} and the release of \textit{VQA v1}, an open-access dataset with, as of April 2017, $204,721$ images extracted from MS-COCO~\cite{MSCOCO}, $1,105,904$ free-form and open-ended questions and $11,059,040$ ground-truth answers\footnote{\url{https://visualqa.org/}}~\cite{VQAoriginal}. However, since the formulation of the task, there has been a significant increase of published datasets with varying degrees of difficulty and image-question distribution balancing~\cite{hudson2019gqa, Salewski2022, zhan2020medvqa, vqacpv2}, with some attaining to a specific domain, e.g. medicine~\cite{zhan2020medvqa, gong2021sysu}, while others focusing on stressing models' learning process through out-of-distribution configurations for the data~\cite{vqacpv2, kervadec2021roses}.

Approaches to achieve a high level of accuracy in VQA have equally varied through time. As an initial attempt, \cite{VQAoriginal} achieves $57.75\%$ in their own dataset using a pre-trained VGGNet for image processing and a deep LSTM for question processing in a supervised learning scheme. Since then, many works have focused on more complex architectures and sophisticated modules for VQA, such as bias-mitigation architectural branches~\cite{yang2020learning}, trilinear transformers~\cite{wang2021mirtt}, attention mechanisms~\cite{farinhas2021multimodal}, external knowledge incorportation~\cite{chen2021zero}, etc. To the best of our knowledge, since the release of \textit{VQA v2}, a bigger, better distributed dataset following the same principles from its predecessor~\cite{vqav2}, the state-of-the-art for its \textit{test-std} split consists of a large scale pre-trained Multiway Transformer applied to both visual and textual feature extraction, achieving $84.03\%$~\cite{beit-sota}, as will be further explored in section \ref{ss.sota} of the present survey.

According to \cite{VQAoriginal}, for a machine learning model to be successful in the task of open-ended VQA, it needs a list of abilities to correctly handle the image-question input: fine-grained image recognition, object detection, spatial awareness, action recognition and knowledge-based reasoning. Answering questions that are simple to humans, such as the ones exhibited in Figure 1, requires a lot of comprehension from the machine side and a capacity to reason based solely on the image and the question prompt. Since this field of research is under constant transformation and evolution, this survey intends to review recently developed works, providing an explanation to how they approach the task.

\section{Known Approaches}
\label{s.known_approaches}
{In this section, we review the works attained through our selection method. We provide an explanation on their methodologies for each paper, the datasets and metrics used, as well as a discussion of shortcomings and open problems wherever applicable.}

\subsection{Learning Content and Context with Language Bias for Visual Question Answering}

{In \cite{yang2020learning}, the authors discuss language bias in the context of VQA. Models can capture these biases during training and repeat them, leading to scenarios where for a certain type of question, the model tends to output the same answer without paying attention to the image's visual content. The authors stress that while several works attempt to overcome this problem by reducing language bias, this approach leverages the model's capacity for understanding context without underestimating the content of an individual instance. Therefore, they propose a novel learning strategy which consists of jointly learning local key content and global effective context, with a novel loss function to reduce the impact of language bias while retaining its benefits.

To embed the visual features, a pretrained Faster RCNN network~\cite{Faster_RCNN} is applied, and for the linguistic features of the question, a Glove embedding + a Gated Recurrent Unit (GRU)~\cite{gru-original}. Firstly, a bias term $b_i$ is calculated for each image-question pair, in which a probability for every possible answer is retrieved considering only the question's type. The Content branch uses an ensemble of Up-Down models for classical VQA learning without language bias, in which its loss term $\mathcal{L}_{cn}$ is negatively affected by the above-mentioned bias term. Meanwhile, the Context branch consists of two neural networks, projecting the visual and linguistic features to a common space, which are fused and fed to a classifier. In this branch, where biases are considered important, $b_i$ is transformed into a binary vector with the same dimension as the branch's output, and used to positively affect its loss term $\mathcal{L}_{cx}$. The outcomes of both branches are fused by element-wise product and the final-target evaluation is done, with loss term $\mathcal{L}_p$. In the end, the novel loss proposed is a combination of all the calculated losses, defined as $\mathcal{L}_{ccb} = \mathcal{L}_{cn} + \mathcal{L}_{cx} + \mathcal{L}_p$.

In this manner, the authors intend to alleviate VQA models' preference for language priors when answering image-question pairs while retaining its beneficial impact on decisions. Indeed, in their tests using VQA v2~\cite{vqav2}, a rather biased dataset, alongside VQA-CP v2~\cite{vqacpv2}, a benchmark dataset for such biases, the authors were capable of reducing the gap between the overall evaluation of both datasets with their model, surpassing other state-of-the-art methodologies. In a qualitative evaluation, they demonstrated that their novel learning strategy leads models to pay attention to more important areas of the image than its counterparts, an Up-Down and an Up-Down + LMH.}

\subsection{Beyond Accuracy: A Consolidated Tool for Visual Question Answering Benchmarking}

{In \cite{vath2021beyond}, the authors propose a browser-based benchmarking tool for VQA models. They argue that most datasets contain many biases that models can exploit to achieve higher accuracies, and therefore to consider only one "number" (metric) as indicative of a models' ability to answer arbitrary, general questions is misleading. To overcome this issue, a digital tool for analysis, evaluation and testing of models and datasets is necessary. The proposed system focuses on four core concepts: realism, in testing the robustness of models from input perturbations; generalizability, in providing specific datasets to benchmark different properties of each model; explainability, in calculating models' uncertainty and biases besides accuracy to give insight on their behaviors; interactivity, in allowing the end-user to select which metrics they want to analyze and the data-scope they wish to measure.

Six datasets are included in the digital tool to benchmark models: VQA-V2~\cite{vqav2}, the most common baseline for most model comparisons; GQA~\cite{hudson2019gqa}, which is designed to test models' visual reasoning capabilities; GQA-OOD~\cite{kervadec2021roses}, a split of the GQA dataset in which evaluation is performed on out-of-distribution (OOD) data; CLEVR~\cite{clevr}, a synthetic dataset made to test models' capacities of abstract visual reasoning regarding spatial relations, colors, etc; OK-VQA~\cite{okvqa}, which requires outside knowledge for questions to be correctly answered; and TextVQA~\cite{TextVQA}, a dataset in which reasoning of text inside images is necessary to respond the questions.

Many metrics are evaluated, such as the official VQA-V2 accuracy, modality bias, robustness to noise - in which noise can be added to original images to measure its impact on model behavior -, robustness to adversarial questions, and uncertainty. The tool also provides different ways to view models performance: globally, in which evaluations are performed on entire datasets; metrical, where graphs on all metrics evaluated in the model are presented to the user; filtering, for understanding models' suspicious behavior on specific questions; and sample, where a single data instance can be analyzed through the model.

To test the proposed tool, four widely-known VQA models are used: BAN~\cite{ban-original}, MCAN~\cite{mcan-original}, MMNASNET~\cite{mmnas-original}, and MDETR~\cite{mdetr-original}. The authors present results regarding the calculation of many metrics as well as a study of these models' generalization capabilities on datasets they were not trained on. The results indicate that models cannot generalize well to data that require different answer spaces, leading to the authors' suggestion of developing VQA systems that are able to generate answers instead of interpreting them as multiple-choice. In the end, to the best of their knowledge, the contribution this system represents is the first of its kind for the VQA domain. In the future, they pretend to work on adding new datasets, new metrics for model evaluation, and on providing more detailed analysis, as well as increasing interactivity for the experiments.
}

\subsection{Multimodal Continuous Visual Attention Mechanisms}

{In \cite{farinhas2021multimodal}, the authors propose a novel attention mechanism for the VQA task that more closely resembles human attention for answering questions based on images. They emphasize that while discrete attention mechanisms are very flexible, such flexibility can result in lack of focus through scattered attention maps. Following the work of Martins et al.~\cite{martins2020sparse} where a continuous unimodal attention mechanism is introduced, the authors of \cite{farinhas2021multimodal} intend to develop more robust attention mappings owing to the fact that some images may require model focus with complex or non-contiguous patches. Therefore, they propose to extend the unimodal attention by developing a multimodal counterpart.

From \cite{martins2020sparse}, instead of feature vectors, images are represented as continuous feature \textit{functions} $V : \mathbb{R}^2 \rightarrow \mathbb{R}^D$, and attention scores are represented by a quadratic score function $f : \mathbb{R}^2 \rightarrow \mathbb{R}$ where relevance is directed towards a single point in the image, with its influence over the image resembling an ellipse. This score function is mapped to a probability density $p : \mathbb{R}^2 \rightarrow \mathbb{R}_+$ via a regularized attention mapping, resulting in a Gaussian density $p(\boldsymbol{x}) = \mathcal{N}(\boldsymbol{x}; \boldsymbol{\mu}, \boldsymbol{\Sigma})$ when conjoined with entropy regularization. It follows that, for a single point in the image, the model's attention is redirected towards it through this probability distribution.

To extend this behavior to multimodal attention distributions, the authors of \cite{farinhas2021multimodal} model the probability density as a mixture of unimodal distributions with the formula:
\begin{equation}
	p( \boldsymbol{x} ) = \sum_{k=1}^{K} \pi_k p_k( \boldsymbol{x} )
\end{equation}
where $p_k$ are the individual unimodal distributions and $\pi_k$ are mixing coefficients that define the weight of each component. To learn the parameters $\boldsymbol{\pi} = [ \pi_1, \dots, \pi_k ]^\top$, the Expectation-Maximization (EM) algorithm~\cite{em-algorithm} is used. The optimal $K$ is determined by a model selection method via Bayesian Information Criterion (BIC)~\cite{bic-algorithm} or Minimum Description Length (MDL)~\cite{mdl-algorithm}.

To test their novel attention scheme, the authors used the VQA-v2~\cite{vqav2} and VQA-HAT~\cite{vqa-hat} datasets and adapted an MCAN~\cite{mcan-original} model, representing the image inputs through a pretrained ResNet50~\cite{he2016residual}, and the questions through a 300-dimensional GloVe~\cite{glove} embedding. For reference, they compare the three attention methods described in this paper - discrete, unimodal continuous and multimodal continuous - and the results indicate that for the \textit{test-dev} and \textit{test-standard} splits of VQA-v2, the discrete attention performed better, albeit the accuracy differences between attention schemes were very low. However, the authors argue that their proposition really excels at mimicking human attention, reporting the Jensen-Shannon (JS) divergence~\cite{js-divergence} for all three attention schemes on the VQA-HAT ground-truth annotations: the discrete obtained 0.64, while the unimodal achieved 0.59 and the multimodal, 0.54.}

\subsection{Roses are Red, Violets are Blue... But Should VQA Expect Them To?}
\label{resume.roses}

{In \cite{kervadec2021roses}, the problem of VQA models learning to exploit language biases in datasets in order to achieve high performance is discussed. The authors defend that measuring simple in-domain accuracy is misleading and does not provide clear insight on the model's generalization capabilities. Therefore, they argue that a new evaluation protocol is necessary to adequately benchmark VQA models and specifically to understand how well they generalize to Out-Of-Distribution (OOD) settings, a way to infer if the model is in fact developing high-level reasoning on the image-question prompt, the main focus of the VQA task.

To achieve this, a new benchmark dataset based on the GQA 1.2~\cite{hudson2019gqa} is proposed: GQA-OOD. It consists of nearly 54,000 questions divided in 4,000 groups, all with close to 10,000 images. The purpose of this dataset is to evaluate models on OOD situations by using infrequent context for image-question pairs on its test set. This is done by separating the questions into groups according to context, extracting the most imbalanced groups regarding their answer distribution and selecting samples from them. Three new metrics are also defined for such benchmark. They are: \textit{acc-tail}, that measures the accuracy on OOD samples; \textit{acc-head}, that measures in-distribution samples; and \textit{acc-all}, the classical in-domain accuracy.

The authors experiment with a variety of models and using different bias-reduction techniques. They find out that models tend to fail in generalizing to infrequent association of concepts (when, for example, a "rose" is not "red"), and the bias-reduction techniques used are all incapable of improving the models' \textit{acc-tail}. In the end, the authors present a comparison between GQA-OOD and other benchmark datasets - GQA~\cite{hudson2019gqa}, VQA-v2~\cite{vqav2}, VQA-CP2~\cite{vqacpv2} -, highlighting improvements and shortcomings of both.

The work documented in \cite{kervadec2021roses} presents a novel benchmark dataset for explicitly highlighting models' tendencies to rely on biases for high performance in the VQA task. They also propose new metrics for evaluating specifically OOD and in-distribution ones. Finally, an extensive study on state-of-the-art models is conducted and their performance is measured in the proposed benchmark, arriving at the understanding that such models are bias-dependent in order to achieve high results.
}

\subsection{Unshuffling Data for Improved Generalization in Visual Question Answering}

{In this paper~\cite{teney2020unshuffling}, the authors address the problem of VQA dataset biases and how it affects models' learning. The task of Visual Question Answering contains many datasets that are known to have powerful biases in their answer distribution, leading the recent literature to apply quite some focus to this issue. However, instead of developing a novel bias-reduction module, the authors of \cite{teney2020unshuffling} propose to separate the training samples into well-chosen partitions to eliminate spurious correlations in the data while maintaining stable ones.

Most VQA models have a feature extraction part alongside a linear classifier. The method proposed consists of optimizing a different copy of the model for each training environment (data partition) such that the feature extraction weights are shared across all copies, but the linear classifiers are not. This leads the image recognition to be the same across all environments, but the reasoning behind it to differ depending on the partition. Lastly, a variance regularizer is applied to each linear classifier's weight to encourage convergence. At test time, the authors use the arithmetic mean of these weights to compute the models' predictions.

The authors experiment their method in three ways: using VQA-CP~\cite{vqacpv2}; using GQA~\cite{hudson2019gqa}; and using VQA v2~\cite{vqav2} alongside Visual Genome QA~\cite{Visual_Genome}. In the first experiment, a comparison with other state-of-the-art models is done, and the partitions selected are the different question types (\textit{yes/no}, \textit{numbers}, and \textit{others}). They find that, while Clark et. al's methodology~\cite{clark2019dont} performs better on \textit{yes/no} and \textit{numbers} categories (as well as \textit{overall}), the authors' achieve better results on the \textit{others} categories. Also, a constant increase of around 3\% is seen from their baseline model to their proposed method, even when using an ensemble for the final results.

In the second experiment, the training environments are selected among equivalent questions, i.e. two questions that query the same information, but are written differently. The authors report graphically the results from this experiment showing clearly that the proposed methodology outperforms the baseline model even with data augmentation - arguing, however, that such augmentation changes the answer distribution and therefore makes the training set incompatible with the test set in this sense. In the final experiment, two environments are selected - one containing VQA v2 and the other VG-QA - to test whether this work's contribution is more impactful than simply aggregating the two datasets together. Indeed, it is reported that the proposed method achieves a slightly better performance on the VQA v2 dataset than simply training the baseline model on both datasets at once.}

\subsection{Structured Multimodal Attentions for TextVQA}

{The authors of \cite{gao2021structured} initially argue that most applications of VQA in assisting visually-impaired people must correctly understand text in images. However, models that tackle such task (usually through the TextVQA dataset~\cite{TextVQA}) suffer from poor text reading ability and lack of textual-visual reasoning capacity. To this end, a novel end-to-end Structured Multimodal Attention Neural Network is proposed. The authors use a structural graph representation to encode object and text relationships in the image, designing a multimodal graph attention network to reason over the queries. This attention is fed to a global-local attentional answering module to produce an answer iteratively (instead of discriminatively). The proposed system consists of three main modules: a Question Self-Attention Module (QSAM), a Question Conditioned Graph Attention Module (QCGAM) and a Global-Local Attentional Answering Module (GLAAM).
	
The focus of the QSAM is to embed the question features that query the image. Given a question with $T$ words, a pre-trained BERT~\cite{bert-original} is used to obtain the question features $\{x_t\}^T_{t=1}$. Since six types of representation are needed to construct a meaningful graph - object (\textbf{o}) and text (\textbf{t}) nodes; object-object (\textbf{oo}), object-text (\textbf{ot}), text-object (\textbf{to}), and text-text (\textbf{tt}) edges -, a linear projection of the features through a Multilayer Perceptron (MLP) is done, and the final decomposed questions features $\boldsymbol{s}^m$ is computed in an element-wise product between the initial features and the linear projections, where $m = \{o, oo, ot, t, tt, to\}$.

The QCGAM models the question's graph representation and attaches edges to the five closest points of each node. With the decomposed question features $\boldsymbol{s}^m$, a question-conditioned graph attention is achieved. From this, image features (including text in images) are extracted to create learnable node representations, as well as attention weights constructed with the linear projections of the features in QSAM, which result in node attention weights. Finally, the node representations and their attention weights are combined via element-wise product, resulting in question-conditioned features for the next module, GLAAM.

This module follows the premise of M4C~\cite{hu2020iterative}, but instead of the first decoding step containg the token \texttt{<begin>}, it contains the summarized global features of the question, object and texts, along with local Optical Character Recognition (OCR) embeddings. These global features are obtained from concatenating the decomposed question features $\bar{s}^o = [s^o ; s^{oo} ; s^{ot}]$ and $\bar{s}^t = [s^t ; s^{tt} ; s^{to}]$, where the authors use "$;$" to denote concatenation.

To properly experiment their proposition, the authors publicly provide OCR ground-truth annotations for the TextVQA dataset~\footnote{\url{https://github.com/ChenyuGAO-CS/SMA}}. While not being able to surpass TAP's~\cite{yang2021tap} results on the TextVQA dataset (with and without extra data), the authors' best combination of features (BERT for encoding, SBD-Trans as the OCR system~\cite{sbd-trans-1, sbd-trans-2}, MultiFeats + RecogCNN~\cite{hu2020iterative, sbd-trans-2} for OCR token representation, and a decoder output module) overcomes all other results, achieving 45.51\% accuracy on TextVQA's test set. The authors also provide an ablation study with many different components and a qualitative analysis of their model's reasoning capabilities.}

\subsection{Zero-shot Visual Question Answering Using Knowledge Graph}

{In \cite{chen2021zero}, the authors propose a new system for incorporating external knowledge to a VQA model. They argue that existing methods mostly use a pipeline approach for knowledge matching and extraction, which commonly leads to error cascading and poor performance. Furthermore, the answer bias issue - where an answer did not appear during training - is mostly forgotten in these methods. To solve these issues, the authors propose a zero-shot (ZS) VQA algorithm to use with an external knowledge graph (KG). To properly evaluate the model, they enhance the F-VQA dataset~\cite{wang2016fvqa} to consider unseen answers, developing the ZS-F-VQA dataset.

In order to encode the relationships present in the image-question pair $(i, q)$, the authors define three different embedding spaces. The knowledge space is the concomitant embedding of $(i, q)$ and the answer $a$ to the same feature space. The semantic space focuses on language information in $(i, q)$, in which $q$ weighs more, working as a guidance to project triplet relations in the KG. The object space acts as a support entity classifier to avoid the direct learning of complex knowledge. An answer masking step is also implemented, receiving the embeddings of the semantic and object space to arrive at a reasonable answer. This is done by collecting the most probable answers and to the given $(i, q)$ query and continuing only with the ones somewhat related in the KG. A similarity score is calculated with the hyperparameter $s$ - the masking part of the system - to give advantage to such answers.

Testing in the F-VQA~\cite{wang2016fvqa} dataset, the authors' method overcome state-of-the-art (SOTA) results in \textit{Hit@1}, \textit{Hit@3} and \textit{Hit@10}, alongside the Mean Rank (MR) and Mean Reciprocal Rank (MRR) metrics. In their novel ZS-F-VQA dataset, the same occurs: their method, with varying hyperparameters, achieves better performance than all SOTA models considered. During analysis of their model's components, the authors find that GloVe~\cite{glove} embeddings worked best for their task. They also provide a qualitative analysis of their model's predictions.}

\subsection{How Transferable are Reasoning Patterns in VQA?}

{In \cite{kervadec2021transferable}, the authors pose the question of whether the inclusion of additional bias-reduction branches in models, or the manual removal of spurious correlations between samples in training data, is necessary for a VQA model to perform without relying greatly on such statistical shortcuts. To this end, they conjecture that visual uncertainty is the main bottleneck in the reasoning pipeline, proposing a study with a perfect-sighted visual oracle - a model that replaces object detection output, usually from a pre-trained visual feature extractor, with ground-truth object annotations directly from the dataset used for evaluation -, the transfer of its learned correlations to deployable models - i.e., with common object detection output -, and an analysis of these models' performances on out-of-distribution settings.

Initially, the authors train an oracle model alongside a classical one, both using a tiny-LXMERT~\cite{lxmert} backbone, and evaluate their performances on the GQA-OOD~\cite{kervadec2021roses} dataset (described in section \ref{resume.roses} of this survey), focusing specially on the out-of-distribution, rare data samples. Although trivially the oracle model achieves incredibly high accuracy (close to 90\%) on common question types, while the classical model comes close to 60\%, the authors shed light on the fact that, on infrequent questions, the visual oracle maintains a high accuracy (around 80\%) and the classical does not (22\%). As a result of this, the authors conjecture that the visual oracle comes closer to a real reasoning process, through the manipulation of words and objects, rather than simply capturing statistical shortcuts (unfeasible in out-of-distribution scenarios), while admitting that no formal proof of such conjecture exists.

A study concerning the transformer architecture's visio-linguistic attention heads energy activation (following the footsteps of \cite{hopfield}) is provided using a $k$-scheme: for a given head and a given sample, the amount of tokens $k$ needed to activate 90\% of the distribution energy is calculated. This is used to indicate attention heads that need a large amount of information from the image or the question in order to function properly. The authors also provide a study of the attention heads activation for specific function types - "what color is...", "where is...", etc -, and notice that linguistic attention heads are more dependent on the function type than most visual heads. All energy visualizations are made through an online tool that the authors released~\footnote{\url{https://reasoningpatterns.github.io}}.
	
Finally, the authors develop the transferring of reasoning patterns from the oracle model to a classic one. They first train the oracle model, initialize the "classical" with the oracle's parameters, continue training BERT-like/LXMERT's objectives, and fine-tune on the evaluation dataset. The architecture used is the tiny-LXMERT~\cite{lxmert} and the datasets used for evaluation are GQA~\cite{hudson2019gqa}, GQA-OOD~\cite{kervadec2021roses} and VQAv2~\cite{vqav2}. They report an increase in comparisons between their strategy with the oracle model in regards to the baseline, with and without LXMERT post-training: a 4.4 accuracy points increase in GQA (without LXMERT post-training) and a 1 point increase in the same dataset (with LXMERT post-training). They also conduce an ablation study with this transferring pipeline and find that transferring the oracle's parameters alongside the LXMERT post-training increases 4.6 accuracy points w.r.t. the baseline. Finally, comparing different SOTA models, their strategy achieves higher accuracy than the other best performing model, MCAN-6~\cite{mcan-original}; an accuracy increase is also seen when applying their method to the simple tiny-LXMERT, but the non-tiny version of the architecture still achieves higher results (around 1.5 points higher in both GQA and GQA-OOD).}

\subsection{Medical Visual Question Answering via Conditional Reasoning}

{Medical Visual Question Answering (Med-VQA) is a hotly debated topic in the area due to its propensity for more accurate medical diagnoses and also for the ease to reaching these conclusions, even bringing new perspectives on the same problem.
	
	In this paper~\cite{zhan2020medvqa}, the authors proposed a novel conditional reasoning framework, using de module proposed, “a question-conditioned reasoning (QCR) module to guide the modulation of multimodal fusion features”, adding a question-conditioned module to the reasoning module, allows the MED-VQA system to find the correct answer and beat open-ended questions considered worst to perform. The reasoning module extracts “task information” from the question and guides the modulation of learning a high-level of reasoning skills by the imposition of the “importance selection over the fusion features”.
	
	This work outperforms the state-of-the-art in VQA-RAD dataset and gives evaluations. It is easy to apply like a “plug and play” and can increase models up to 10.8\%.

}

\subsection{IQ-VQA: Intelligent Visual Question Answering}

{Consistency and robustness are a problem in VQA models based, and to contour this problem the paper \cite{vatsal2020iqvqa} debated the theme.The authors proposed in this paper a “model-independent cyclic framework”, the objective is to increase consistency and robustness. Furthermore, a novel implication generator is proposed and can generate “implied questions from any question-answer pair”, in general, the model needed to answer the question proposed and then, the question created by that answer. The last contribution of this work is a new “human annotated VQA-Implications dataset”, that consists of 30K questions and is separated into 3 types of implications: Logical Equivalence, Necessary Condition and Mutual Exclusion.  
	
	In this work, the authors trained the model to:  answer the original question, generate an implication based on the answer and learn to answer the generated implication correctly.
	
	This framework improves the consistency of VQA models by almost 15\% in the rule-based dataset, 7\% on VQA-Implications dataset and robustness by 2\% without degrading their performance.

}

\subsection{Separating Skills and Concepts for Novel Visual Question Answering}

{Although the advances of VQA, still have problems in the models to discuss, one of them is the generalization to out-of-distribution data, in this paper~\cite{spencer2021skill} the authors present a new method based on separating ‘skill’ and ‘concepts’.
	
	To contour the problem mentioned, the authors proposed to separate the measure generalization into “skills and concepts”, the explanations of them are “ ‘Skills’ are visual tasks, such as counting or attribute recognition, and are applied to ‘concepts’ mentioned in the question, such as objects and people”. The authors present a novel method for learning to compose them, and experiments prove the effectiveness of this method. The main dataset used in this paper is VQA v2, and the results can achieve state-of-the-art performance as well from generalizing unlabelled data.}

\subsection{Debiased Visual Question Answering from Feature and Sample Perspectives}

{Many VQA models are perturbed by biases and just capture them, prejudicing the evaluations method and the AI to show an capability of real reasoning.
	
	In this paper~\cite{wen2021debiased}, the authors created the D-VQA method which decrease the negative effect of both language and vision biases in the feature and sample perspective. Building a question-to-answer and a vision-to-answer feature, this method is capable of capturing language and vision biases. Furthermore, the authors used two unimodal bias detection modules to explicitly recognise and remove them. To assist the models in their training, two negative samples were built.
	
	D-VQA achieves incredible results in VQA-CP v2 (test set) and VQA v2 (validation), the first annotation that is notable, the authmentation-based methods surpassed non-augmentation based methods, demonstrating that reducing bias with focous on data is more effective. The second annotation, says about the performance of D-VQA, surpassing CF-VQA \cite{niu2021counterfactual}, SSL-VQA \cite{zhu2020overcoming} and Mutant\cite{gokhale2020mutant} in the same dataset, the authors adds that they constructed a negative sample based on available samples, without using additional annotations. The third annotation shows the effectivenness of D-VQA, while the most methods in VQA v2 performs worse than UpDn, D-VQA surpasses, besides, D-VQA perform better in IID (VQA v2) and ODD (VQA-CP v2) datasets.}

%
%
%
%

\subsection{Graphhopper: Multi-hop Scene Graph Reasoning for Visual Question Answering}

{The ability to associate semantic and linguistic understanding of objects in an image to get an answer is something to concern in Visual Question Answering (VQA) when we talk about the answer of free-form questions.
	
	To approach that task, the authors created Graphhopper~\cite{rajar2021graphhopper} the first VQA method that employs reinforcement learning for multi-hop reasoning on scene graphs. The framework includes computer vision, knowledge graph reasoning and natural language processing techniques. These techniques extract information of the image, focusing on objects and getting more context identifying what compound that objects, to associate and answer a question with more accuracy. According with authors, an agent is trained to extract conclusive reasoning paths from scene graphs, navigating in a multi-hop to reach the objective.
	
	The authors conducted experimental studies on the GQA \cite{hudson2019gqa} dataset, in manually curated scene graphs, Graphhopper reaches human performance. Besides, Grapphoper outperformed the Neural State Machine (NMS)\cite{NSM_graphhopper}, that is a state-of-the-art in scene graph reasoning. For future work, the authors plan to combine scene graphs with common sense knowledge graphs. The framework outperforms Neural State Machine (NSM), a similar work.  }

\subsection{Psycholinguistics meets Continual Learning: Measuring Catastrophic Forgetting in Visual Question Answering}

{When supervised machine learning models are incapable of continuously learning new tasks, the authors call that ‘Catastrophic Forgetting’, and to contour this problem, some studies used ‘Continual Learning’\cite{continual_learning_original}. In this paper~\cite{claudio2019psycholinguistics}, the authors studied that in Visual Question Answering (VQA) with psycholinguistic evidence.
	
	The authors used characteristics that are known to be learned by children to build two tasks, both tasks are assigned two types of questions, first Wh-questions (questions about the attribute of an objet) and then Y/N-questions (questions that compare objects with respect to an attribute) using the CLEVR dataset. Wh-question results in better accuracy than Y/N-questions and reveals it’s more important to learn them first, to get facilitates in continual learning, exactly like psycholinguistics. The dataset used is CLEVR\cite{clevr}.
	
	The paper is well structured and easy-to-understand with various examples about the theme and reaches great results, allowing new studies in the area using new datasets. 
}

\subsection{LaTr: Layout-Aware Transformer for Scene-Text VQA}
{In this paper, a new transformer based architecture to solve ST-VQA (Scene Text Visual Question Answering) \cite{STVQA_original} was created, named LaTr (Layout-Aware Transformer). ST-VQA is the task that will join the basis of VQA, with image text recognition to answer a previous question. Reaching a new state-of-the-art, the authors present a new symbiosis between Scanned Documents and scene text, being the first to propose a pre-training in these documents instead of natural images. These techniques improve robustness in common OCR errors, and take off the needs of an External Object Detector. Besides, in OCR, the authors used Rosetta-en \cite{Rosetta_OCR} and Amazon-OCR \cite{Amazon-OCR}  in different tasks, to complement, the state-of-the-art was reached with both.
	
	More technically, this new architecture has a language model based on Text-to-Text transformer and uses the base of T5 \cite{T5} because of its extensive pre-training data using Common Crawl, that is 750GB of cleaned English text data. To complement, a pre-training with C4 is done with a denoising task. Furthermore, to encourage the model to fully use the layout information, which is a problem cited, the authors give access to the rough location of the masked tokens. In this part, cross-entropy loss was used.
	
	Also, this work uses ViT \cite{ViT} to extract visual features pre-trained and fine-tuning on ImageNet \cite{ImageNet}, and profes that are better than FRCNN \cite{Faster_RCNN}  in a scalable way. The datasets used in this work were TextVQA \cite{TextVQA} to fine-tuning in the language module, ST-VQA \cite{STVQA_original} to fine-tuning and the newest the paper presents, the pre-train into Scanned Documents, are used IDL.
	
	Therefore, authors conclude with an important idea to follow in ST-VQA task, "language and layout are essential". Furthermore, the authors add an appendix with examples and more technical reports, with the complete architecture of the model, and give more context about the bias problem, related to the non-captioning layout in all tasks and give a work to the future to better explore this question. }

\subsection{MUST-VQA: MUltilingual Scene-text VQA}
{The authors of \cite{MUSTVQA} introduce a new task more generalized of Scene Text VQA (ST-VQA) \cite{STVQA_original} named Multilingual ST-VQA (MUST-VQA), this new task handle with languages other than English, specifically, the authors propose a initial study with questions in English, Spanish, Catalan, Italian, Chinese and Greek, creating a framework to solve this new task.

Furthermore, due to the difficulty of obtaining new data and having only English datasets, authors separated into constrained and unconstrained settings, unconstrained setting refers to a STVQA model that accepts questions, read/analyze and produce answers in any language. Knowing the limitations into constrained settings, a division into IID and Zero-shot fashion was proposed, that is, an evaluation of the model with languages that is trained and languages it has never seen.

Technically, the framework is transformer-based, meaning that it has a visual and textual encoder. In the Visual Encoder, to obtain salient regions of an image, a Faster R-CNN \cite{Faster_RCNN} pre-trained with ImageNet \cite{ImageNet}, and later, fine-tuned with Visual Genome \cite{Visual_Genome}. In Textual Encoder according to the authors, the questions are embedded through a given encoder to obtain a set of features to be used as a representation to be later fed into a transformer-based model. Many textual encoders are used, they are: Byte Pair Encoding (BPEmb) \cite{BPEmb}, Bidirectional Encoder Representations from Transformers (BERT) \cite{bert-original}, Multilingual-BERT (M-BERT), Text-to-Text Transfer Transformer (T5) \cite{T5}, Multilingual-T5 \cite{mT5}.

Furthermore, in this paper, the models M4C \cite{hu2020iterative} and LaTr \cite{LaTr} are adapted to the task of MUST-VQA, the new adapted model names M5C-mBERT, that substitute FastText \cite{FastText} by OCR tokens, and replaced PHOC \cite{PHOC} representation to accept new scripts, finally are employed BPEmb. In LaTr, now named mLaTr, the T5 encoder-decoder are replaced by mT5, which are fine-tuned only the text pre-trained multi-lingual Language Model with the multimodal information.

In Experiments section, the authors used ST-VQA and TextVQA \cite{TextVQA} to benchmark, since the dataset of MUST-VQA consists into a ML-STVQA and ML-TextVQA, which are obtained before translate the previous dataset with Google-translate-API, furthermore, Rosetta-OCR \cite{Rosetta_OCR} and Microsoft-OCR are used to detection. In TextVQA, basically, all new models outperform the original, the best results occur when models use Microsoft-OCR, in the different languages that are proposed, not excluding English, in addition the results with zero-shot are more discrepant in this task. The results in ST-VQA follow the TextVQA. Reported results show that visual features do not increase accuracy in general. In all experiments, first are trained with the IID language, to finally test in zero-shot. Finally, Adam \cite{Adam_original} optimizer was used in all M4C-based methods and AdamW \cite{AdamW} in all T5-based models, accuracy is the main metric used in all works.

The authors still bring a result table related to different machine translation models into the best model tested, mLaTr, these machine translation models namely OPUS, M2M-100 and mBART, surprisingly, the results are coherent with the original translation model.

Finally, in future work, the authors express the need for approaches into the answer module to different languages, since this paper only works with the question.}

\subsection{MIRTT: Learning Multimodal Interaction Representations from Trilinear Transformers for Visual Question Answering}

{General VQA solutions only use the image and the question as input while trilinear models benefit from the fact that answers contain relevant information. However, trilinear models do not consider the intra-modality information. In \cite{wang2021mirtt}, the authors try to fill this gap in the literature by proposing a trilinear modalities interaction framework called MIRTT (Learning Multimodal Interaction Representations from Trilinears Transformers).

The main contributions of this work are proposing a framework based on the TrI-Att and Self-Att mechanisms and a two-stage workflow to simplify Free-Form Open-ended VQA (FFOE) into Multiple Choice (MC) VQA.

For text extraction from questions and answers a BERT mode was fine tuned while a Fast R-CNN detector was used without fine-tuning. They present the experiment results of MiRTT on both FFOE VQA and MC VQA. The metrics evaluated are an accuracy-based evaluation metric (ACC) and Acc-MC.

Besides higher accuracy values on trilinear approach. The MIRTT also had good results when being applied as a second stage after bilinear methods as backbones.
The authors present an ablation study on the advantages of MIRTT components proving their effectiveness.}

\subsection{Contrastive Pre-training and Representation Distillation for Medical Visual Question Answering Based on Radiology Images}

{The work \cite{liu2021constrative} aims to create a dataset from an unlabeled set of radiography images. Given or unlabeled set of images they train a teacher model, that uses augmented images of three
body regions (brain, chest, abdomen) in a self supervised manner to learn differences among these regions. They applied the Momentum Contrast, a self-supervised contrastive learning method.

The inter-regions features are obtained through the training of a lightweight student model. The images are grouped by region in open source databases, so the class labels for each image are already known. The discriminative features are obtained with the classification
loss with a balancing parameter.

They ran experiments with VQA-RAD~\cite{zhan2020medvqa} and STAKE datasets. The proposed CPRD framework shows improvements when compared to the MEVF + BAN (5\% and 2\%, respectively). Their results suggest that pre-training on domain specific images is an important step for the VQA problem.}

\subsection{Answering Questions about Data Visualizations using Efficient Bimodal Fusion}

{In the paper \cite{kafle2020answering}, the authors propose a novel CQA algorithm called parallel recurrent fusion of image and language (PreFIL). PReFIL has two parallel Question and Image (Q+I) fusion branches, where each branch takes in question features (from an LSTM) and image features from two locations of a 40-layer DenseNet~\cite{huang2018densely}, i.e. low-level features and high-level features. Each Q+I fusion block concatenates the question features to each element of the convolutional feature map, and then it has a series of 1 x 1 convolutions to create question-specific bimodal embeddings. These embeddings are recurrently aggregated and then fed to a classifier that predicts the answer. Despite being composed of relatively simple elements, PReFIL outperforms more complex methods that use RNs and attention mechanisms.

The authors critically review the publicly available CQA datasets, FigureQA~\cite{kahou2018figureqa} and DVQA~\cite{kafle2018dvqa}, outlining their strengths and weaknesses, by doing extensive experiments. For DVQA, an additional fourth OCR-integration component is required.

PReFIL surpassed prior state-of-the-art methods for both DVQA and FigureQA. While PReFIL exceeded the human baseline for FigureQA, results are more nuanced for DVQA due to OCR model variations. All OCR versions exceeded the human baseline for structure questions, but only PReFIL using oracle OCR exceeded humans across all question types. We found that better OCR methods led to better results for DVQA. Future developments in OCR technology would likely improve PReFIL further. The strong results in this paper suggest that the community is ready for more difficult CQA (Community Question Answering) datasets.}

\subsection{Coarse-to-Fine Reasoning for Visual Question Answering}

{To bridge the semantic gap between images and questions in VQA, the authors~\cite{nguyen2022coarse} introduce a new framework that focuses on reasoning the visual contents in the image and the semantic clues in the question in a coarse-to-fine manner.
	
The Coarse-to-Fine Reasoning (CFR) framework takes an image and a question as inputs. The image is passed through the Image Embedding module to extract the region of interest (RoI) features and visual predicates. The question is processed in the Question Embedding module to extract the question features and question predicates. The predicates are keywords about objects, relations, or attributes of the image/question. To effectively map the visual modality and language modality, the authors jointly learn their features, as well as their predicates in the Coarse-to-Fine Reasoning module.

To effectively map the information of the question to the visual information in the image, the Coarse-to-Fine Reasoning module utilizes three steps: Information Filtering, Multimodal Learning, and Semantic Reasoning. The Information Filtering aims to filter out unnecessary visual information from the image based on the predicates. The Multimodal Learning module learns the semantic mapping between the question and image at coarse-grained and fine-grained levels. Finally, the Semantic Reasoning module combines the output of the multimodal learning step to predict the answer.
They design the Multimodal Learning module to jointly learn the features from the visual and language modalities. Multimodal learning is essential for identifying the correlation between each instance in the image and the question, then identifying which instances in the image are useful for answering the question. In this module, the features are jointly learned at two levels: coarse-grained and fine-grained. The coarse-grained level learns the interaction between question features and image features, while the fine-grained level learns the interaction between filtered information of the image and question obtained from the Information Filtering step.

Then the authors conduct intensive experiments to validate their methods, using three popular datasets: GQA~\cite{hudson2019gqa}, VQAV2~\cite{vqav2}, and Visual7W~\cite{zhu2016visual7w}. The intensively experimental results on three large-scale VQA datasets show that the proposed approach achieves superior accuracy comparing with other state-of-the-art methods. Furthermore, the reasoning framework also provides an explainable way to understand the decision of the deep neural network when predicting the answer.}

\subsection{Question-controlled Text-aware Image Captioning}

{To generate personalized text-aware captions for visually impaired people, the authors~\cite{hu2021question} propose a new challenging task named Question-controlled Text-aware Image Captioning (Qc-TextCap). They use questions about scene texts to control text-aware caption generation due to its convenience in interaction. The Qc-TextCap task requires models to comprehend questions, find relevant scene text regions and incorporate answers with an initial caption to produce a final text-aware caption. Based on two existing text-aware captioning datasets, they automatically construct two datasets, ControlTextCaps and ControlVizWiz, to support the task.

They further propose a novel Geometry and Question Aware Model (GQAM) for the  Question-controlled Image Captioning task. GQAM consists of three modules, namely Geometry-informed Visual Encoder, Question-guided Encoder and Multimodal Decoder. The Geometry-informed Visual Encoder fuses object region features and scene text region features with relative geometry information. Question-guided Encoder dynamically selects relevant visual features to encode questions. The Multimodal Decoder takes inputs of visual, question and initial caption features to sequentially generate a text-aware caption for the question.

On both datasets constructed by the authors, ControlTextCaps and ControlVizWiz, GQAM achieves better performance than carefully designed baselines on both captioning and question answering metrics. It is further proved that the model with questions as control signals can generate more informative and diverse captions.}

\subsection{SYSU-HCP at VQA-Med 2021: A Data-centric Model  with Efficient Training Methodology for Medical Visual Question Answering}

{The paper \cite{gong2021sysu} describes the authors contribution to the Visual Question Answering Task in the Medical Domain at ImageCLEF 2021 \cite{ImageCLEF2021}. Since the data from medical VQA is relatively limited comparing to general VQA, they design a data-centric model to collect and make full use of the limited data. Besides, as the questions of VQA-Med 2021 reside in the abnormalities of the medical images, they mainly focus on the design of visual representation to classify the abnormality of the medical images.
	
	To address the issue of data limitation, the authors expand the dataset with the data in the previous VQA-Med competition (i.e., VQA-Med 2019 \cite{ImageCLEFVQA-Med2019} and VQA-Med 2020 \cite{ImageCLEF-VQA-Med2020}). To make full use of the limited data, they apply mixup technology to create more samples. For efficient visual feature extraction, label smoothing is applied to stabilize the training progress. Furthermore, they discover the phenomenon that hard samples restrict the performance of the model and use a curriculum learning-based loss function to resolve this issue. Last but not least, to prevent the model from the infliction of intrinsic model bias, the authors propose to ensemble different types of models in exchange for higher model accuracy.
	
	As the model unavoidably contains bias, the authors apply multi-architecture ensemble to further improve the model performance. Comparing to the winner of VQA-Med 2020 that takes more than 30 models to an ensemble, their best submission only contains 8 models by taking the advantage of the hierarchically adaptive global average pooling (HAGAP) structure. All backbones are initialized with the ImageNet pre-trained weight.
	
	The VQA-Med competition applies accuracy and BLEU as the evaluation metrics. Accuracy is calculated as the number of correct predicted answers among all answers. BLEU measures the similarity between the predicted answers and ground truth answers. The proposed method achieves 1st place in the competition with 0.382 in accuracy and 0.416 in BLEU.}

\subsection{VisQA: X-raying Vision and Language Reasoning in Transformers}

{Visual Question Answering systems target answering open-ended textual questions given input images. They are a testbed for learning high-level reasoning with a primary use in Human-Computer Interaction, for instance assistance for the visually impaired. The work was primarily motivated by growing concerns in the field over bias exploitation of models trained in large-scale settings, in particular when trained on very broad problems like vision and language reasoning.

The authors \cite{jaunet2021visqa} introduce VisQA, an instance-based visual analytics tool designed to help domain experts, referred to as model builders, investigate how information flows in a neural model and how the model relates different items of interest to each other in vision and language reasoning. The tool exposes the key element of state-of-the-art neural models as attention maps in transformers. VisQA implements attention heads and interactive heatmaps visualizations, and other interactions such as free-form question and pruning mechanisms, to assess whether a model resorts to reasoning or bias exploitation when answering questions.

The GQA~\cite{hudson2019gqa} standard dataset provides question/image pairs along with their answers, the ground truth of bounding boxes, and semantic descriptions of questions. By default, VISQA provides around 1500 question/image pairs, but as images are loaded progressively when users request it, such a quantity can be increased without affecting performances. The models have been trained on a significantly larger amount of training data (about 9M image/sentence pairs), which is different from the validation data on which the performance is evaluated. The user interface of VISQA is implemented using D3~\cite{6064996}, and directly interacts with transformer models implemented in Pytorch~\cite{NEURIPS2019_bdbca288}, using JSON files through a python Flask server. As the possibility to ask free-form questions offers an infinity of possible combinations, each question/image pair is forwarded through the model in a plug-in fashion, i.e., without altering the model and its performances.

The design process of VisQA was motivated by well-known bias examples from the fields of deep learning and vision-language reasoning and evaluated in two ways. First, as a result of a collaboration of three fields, machine learning, vision and language reasoning, and data analytics, the work lead to a better understanding of bias exploitation of neural models for VQA, which eventually resulted in an impact on its design and training through the proposition of a method for the transfer of reasoning patterns from an oracle model. Second, the authors also report on the design of VisQA, and a user study was conducted with 6 experts with experience in building deep neural networks, who were not involved in the project or the design process of VisQA. VisQA received positive feedback from these experts, who also provided additional qualitative feedback on the nature of the information they extracted on the behavior of different neural models. The quantitative evaluations are encouraging, providing first evidence that human users can obtain indications on the reasoning behavior of a neural network using VISQA, i.e. estimates on whether it correctly predicts an answer, and whether it exploits biases.}

\subsection{Answer Questions with Right Image Regions: A Visual Attention Regularization Approach}

{In the paper \cite{Liu2022}, the authors proposes the application of attention layers in VQA models can lead to problems of prioritization in areas that are not so important for the contextualization of the question-answer, in order to solve this problem, predecessor methods have been doing the task of realigning the attention layer in order to prioritize important areas for the contextualization. However, such a task requires a high cost, mostly because of the high need of human work, which led this study to develop the AttReg, a regularization of the attention layers in order to allow the focus on the most important areas, being a technique that works by finding regions in an image that the main model ignores. Particularly those regions are given low attention weights. AttReg then uses a mask-guided learning approach to regularize those regions and encourage the model to pay more attention to these key regions. The proposed model is flexible and model-agnostic, and can be used into several models created for VQA.

The main idea is to analyze the images with the lowest attention scores, the ignored areas. Masks are applied to the area of ignored objects and the training procedure is performed, serving as a kind of data augmentation. Parameters are updated taking into account normal images and masked images simultaneously. In this way, the model is led to explore other areas of the image, applying a type of regularization in the attention layers.
The use of this regularization method contributed to the improvement of state-of-the-art models, correcting known problems in extracting attributes from images in VQA models.}

\subsection{VLMO Unified Vision-Language Pre-Training with Mixture-of-Modality-Experts}

{The paper \cite{https://doi.org/10.48550/arxiv.2111.02358} proposes an architecture to resolve problems of vision-language classification and image-text retrieval. They introduce Mixture-of-Modality-Experts (MoME) a extension to Mixture-of-Experts (MoE) \cite{jacobs1991adaptive}, MoME is a framework that incorporates multiple modalities, such as audio, text, and image, to improve the performance of a machine learning system. In MoME, each modality is represented by a set of experts that are trained to specialize in different parts of the modality space. Then, the outputs of all modalities experts are combined to form the final prediction.
 
Similar to CLIP\cite{radford2021learning} and ALIGN\cite{zhang2018attention} architectures, VLMo has two encoders for image and text separately. \textcolor{red}{It} presents a new fusion encoder transformer that is responsible for encode the various modalities (images, text and pairs) in a transformer\cite{NIPS2017_3f5ee243} block. Part of the architecture, together with a shared self attention, is responsible for weighing the modalities for the different types of problems. 
 
It has three experts: namely vision expert for image encoding, a language expert for text encoding and a vision-language expert for fusions of text-image. Due to the flexibility of the 3 separate experts, the article proposes an interesting pre-training stage, training image and text experts only with images and texts respectively without the need for data pairs of text-image. Because of this, experts already arrive at the training stage with certain knowledge and this helps the architecture to generalize better. 
 
To validate and analyze the importance of pre-training techniques, the ablation technique was used, which consists of a series of experiments where parts of the architecture are removed in order to assess the importance and impact on the performance of the model. A table containing this experiment is present in the article. The experiments demonstrated that the VLMo reached the state of the art on the VQA and NLVR2 tasks. 
 
One of the main advantages of this architecture is the pre-training possibility, using large amounts of image only and text only data. A disadvantage is the task to organize three sources of data, one with images-only for encoder, other with texts-only for another encoder and a source with image-text pairs for fusion-encoder. }

\subsection{Greedy Gradient Ensemble for Robust Visual Question Answering}

{The article \cite{Han2021} contributes to the reduction of two language bias problems in VQA models. A bias study was carried out and led to the decomposition into two types: distribution bias and shortcut bias. Removal of both types of bias is proposed.
 
In order to solve the problem, part of the data, which are biased, is overfitted in the model, allowing the unbiased data to be learned by the model. Samples with a high chance of fit end up having a low gradient, allowing the model to focus on learning hard samples. Three versions were created for evaluation: distribution bias correction, shortcut bias correction and correction of both.
 
The study was successful in resolving both types of bias, the application of this study in SOTA models can lead to the increase in performance. However, the application of the technique in methods with cross entropy regularization led to a decrease in performance, indicating problems in models with the presence of some type of prior adjustment.}

\subsection{LPF: A Language-Prior Feedback Objective Function for De-Biased Visual Question Answering}

{Looking to improve score at language biased problems, this article \cite{Liang2021} propose an objective function, called Language-Prior Feedback (LFP), with the objective of re-balance the proportion of QA pairs. Imbalanced answer distributions may influence the model to the point of ignoring the information contained in the images and taking into account only the the most frequent answer.

Initially, a model containing questions-only is trained, with the objective of minimizing the cross entropy, later a softmax probability is calculated for the outputs of the question-only model. Losses of each sample are modified, penalizing frequent occurrences and encouraging the learning on other samples. This work had good results outperforming other strong baselines works.
	
The study was validated on the VQA-CP2 and comparisons with SOTAs were made obtaining a improve of 15,85\% over \cite{bottomup}. Another great boost as observed at "Yes/No" questions, obtaining a improve of 43.40\% since this kind of QA is more susceptible to language priors.}

\subsection{Passage Retrieval for Outside-Knowledge Visual Question Answering}

{This article \cite{Qu2021}, the authors presents a novel approach for answering VQA using outside knowledge. The authors propose a passage retrieval technique that leverages external textual sources, such as Wikipedia articles, to provide relevant information. The approach involves encoding the question and the image separately and then retrieving relevant passages from external sources based on the encoded information. The retrieved passages are then used to generate an answer to the question. Authors evaluate the proposed method on two benchmark datasets, OK-VQA and a Wikipedia passage collection with 11 million passages.

The proposed approach addresses a key challenge in visual question answering, which is the limited availability of visual information for answering certain types of questions. By incorporating external textual sources, the method is able to provide additional information that can aid in answering such questions. The authors also highlight the importance of selecting appropriate textual sources for passage retrieval, as using irrelevant sources can lead to inaccurate or incomplete answers. To address this issue, the authors propose a method for selecting relevant sources based on the similarity between the question and the source text.

The study initially conducted a sparse retrieval analysis using the BM25, which is a ranking
function with the objective of estimating the relevance of a query in a specific set of data. A dual
encoder was built in order to combine information generated by the question-answer pair together
with the results obtained using the BM25. In order to improve the search using BM25 information
from the image was extracted capturing object names and image captions.

The experimental results show that the proposed method outperforms existing methods that do not use external knowledge sources. The approach achieves state-of-the-art results on both benchmark datasets, demonstrating its effectiveness in improving the accuracy of visual question answering. The authors conclude that their approach can be used to integrate external knowledge sources in VQA tasks.
}

\subsection{Weakly Supervised Grounding for VQA in Vision-Language Transformers}

{The article \cite{Khan2022WeaklySG} proposes a weakly supervised grounding technique for visual question answering using vision-language transformers. Grounding refers to the process of linking language to the corresponding image regions that it describes. The proposed approach is able to ground answers with no need for precise object localization during training. Instead, it relies on weak supervision, which involves only providing the model with correct answer without specifying the exact image region it corresponds. The model is then trained to learn relevant regions of the image and improve the grounding task.

The approach involves using a vision-language transformer \cite{NIPS2017_3f5ee243} to encode both the image and the question into a joint representation space. The model then attends to relevant image regions based on the question and generates an answer. The answer is then used to compute a gradient that is back propagated to the model's attention weights. The attention weights are then updated to better attend the relevant image regions for the given question.

To do this, image attributes are extracted using a ResNet \cite{he2016residual} and then pass through a Capsule Encoding Layer \cite{NIPS2017_2cad8fa4}, which separates the attributes spatially using n capsules. After this, the capsules are flattened and pass through a fully connected layer to obtain a set of visual attributes. The capsule layer combines the capsules obtained from the encoding layer and aggregates them with texts.

The proposed approach was evaluated in different ways, for GQA \cite{hudson2019gqa} the accuracy was the chosen metric, for measuring grounding task the attention scores obtained on last cross-attention layer was compared to ground truth values obtaining then IOU reported as precision, recall and F1-score. For VQA-HAT, the report is obtainable through of mean rank correlation comparing attention maps generated and the human attention annotations. Overall, the proposed approach provides a promising direction for weakly supervised grounding in VQA using vision-language transformers.

}

\section{Methodology}
\label{s.methodology}

In this section, the most common datasets and metrics used in the reviewed works are described, alongside honorable mentions - datasets not commonly used, but that indicate a recent tendency in the scientific literature -, and how a quantitative comparison was made to understand the state-of-the-art VQA applications in the last few years.

\subsection{Datasets}
\label{ss.datasets}

Developing a robust dataset for VQA is a difficult task, since it involves a large number of annotators and field specialists, specially with the application of deep learning, which usually requires a large amount of data for models to generalize satisfactorily. Because of this, not all datasets used are created entirely from scratch, with some focusing more on developing new splits on already existent data. In this section, we describe the most used datasets of the papers reviewed.

\subsubsection{VQA v1 \& VQA v2}

{VQA v1 was proposed in \cite{VQAoriginal}, this dataset is one of the first ones introduced to measure performance in VQA problems. There are 250k images, which around 205k are from MS COCO \cite{MSCOCO}, 760k questions and around 10M answers. For each image there are three questions with ten open ended answers, meaning that they do not have a fixed set of possible answers. After a few years, the second version was proposed~\cite{vqav2}, with the aim to improve visual importance the V1 version was balanced and more images were collected in a way that each question is associated with at least two images. Since a relation between similar images were created, it become possible to find other question-answer pairs when searching in similar images.}                                                             

\subsubsection{VQA CP v1 \& VQA CP v2}

{The authors proposed in \cite{vqacpv2} the VQA CP v1 and VQA CP v2 versions. Both are created with the objective of reducing bias in answers. The CP v1 consists of VQA v1 but redistributed in train and test sets, while CP v2 is a result of VQA v2 redistribution. The relation of train and set sets is around 60\% in both datasets.}

\subsubsection{Medical VQA Dataset}

{Proposed in \cite{zhan2020medvqa}, this dataset is a manually constructed dataset of medical radiology. All images and the pairs questions-answers were produced by specialists, therefore, making this a high quality dataset. The dataset is composed of 315 radiological images and 3515 visual questions. 
}

\subsubsection{GQA}

{Proposed by \cite{hudson2019gqa}, it includes around 22M questions that have been generated using Visual Genome scene graph structures and functional programs that represent their semantics, combined with 113K images. The dataset also includes a suite of new metrics for evaluating the quality of the answers, measuring semantic understanding, reasoning and consistency. The performance of various models on the GQA dataset has been analyzed, including baseline models and state-of-the-art models, and it has been found that there is significant room for improvement over current approaches.}

\subsection{Metrics}
\label{ss.metrics}

{To evaluate the robustness of VQA models, two trends were most noticeable among the works reviewed: metrics focusing solely on the classification aspect of answers (e.g. accuracy), and metrics that also consider the phrasal structure involved in such answers (e.g. BLEU, CIDEr). In this section, we describe the metrics used for the majority of papers reviewed.}


\subsubsection{Accuracy}

{Also known as VQA Accuracy, the most commonly used metric for evaluating models on the VQA task is the Accuracy. Introduced in \cite{VQAoriginal} for the evaluation of both open-ended and multiple-choice questions, it consists of the following arithmetic:

\begin{equation}
	Acc = \min{ \left( \frac{\text{\# humans that provided that answer }}{3}, 1 \right) }
\end{equation} leading models to achieve the highest accuracy possible if at least three human annotators agreed with its answer.	

Some issues can arise within this metric, such as when annotators do not agree on a certain answer, limiting the highest evaluation that a model's answers can achieve. However, due in parts to its simplicity, the VQA accuracy is the most used metric for evaluating models.}

\subsubsection{BLEU}

{The BiLingual Evaluation Understudy (BLEU)~\cite{papineni2002bleu} is a method for machine translation capable of being used in open-ended answers of the VQA task. It focuses on measuring fluency, adequacy and length for candidate (machine-translated) answers in regards to a reference (human-annotated) one.

It achieves such purpose by calculating a modified precision score for the candidate answers based on an $n$-gram. Firstly, it counts the number of times the candidate $n$-gram appears in any possible reference/ground-truth sentence - here referred to as $C$ -, and then counts the number of times, for each reference sentence, that the same $n$-gram appears - $R_k$, where $k$ indicates which reference sentence it refers to. With this, a count clip is achieved for each $n$-gram by the following equation:

\begin{equation}
	\label{eq.countclip}
	C_{\text{clip}} = \min{ \left( C,  \max{ R } \right) }
\end{equation} where $R = \{R_1, R_2, ..., R_k\}$. The modified precision score is calculated as
\begin{equation}
	\label{eq.modprecision}
	p_n = \frac{ \sum{ C_{\text{clip}} } }{ \sum{ C } }
\end{equation}
To account for the length of words, a brevity penalty $BP$ is calculated. Considering $r_{\text{words}}$ and $c_{\text{words}}$ the number of words in a reference sentence and in a candidate sentence respectively, the $BP$ is achieved by
\begin{equation}
	BP = 
	\begin{cases}
		1 & \text{if } c_{\text{words}} > r_{\text{words}} \\
		e^{\frac{1-r_{\text{words}}}{c_{\text{words}}}} & \text{if } c_{\text{words}} \leq r_{\text{words}}
	\end{cases}
\end{equation}
Finally, the BLEU score is calculated combining all of the aforementioned results:
\begin{equation}
	\text{BLEU} = BP \times \exp \left( \sum_{n=1}^N w_n \log p_n \right)
\end{equation} where $N$ is the number of $n$-grams and $w_n$ the precision weights, usually $w_n = 1/N$.

In summary, (\ref{eq.countclip}) informs how adequate the translation is across all reference answers, while (\ref{eq.modprecision}) measures this adequacy over the entire set of $n$-grams, providing a quantitative way to rank the machine translation.}

\subsubsection{ROUGE-L}

{The Recall-Oriented Understudy for Gisting Evaluation (ROUGE)~\cite{lin2003automatic, rougepackage}, is a machine translation evaluation system composed of 4 different metrics, each with their own intricacies as to how the adequacy of a translation is interpreted. In the recent VQA literature surveyed, the most commonly used ROUGE metric for models' evaluation is ROUGE-L~\cite{rougel}, a longest common subsequence (LCS)-based evaluation procedure.

Following \cite{rougel}, the metric is calculated as follows. Let $c$ be the candidate (machine) translation of size $n$ (i.e. $n$ words) and $R = \{ r_1, r_2, ..., r_u \}$ the set of reference (human-annotated) translations with sizes $m_1, m_2, ..., m_u$ respectively. The $LCS$ between a candidate and a reference translation is the number of words that longest common subsequence contains. Therefore, we calculate two intermediate metrics:
\begin{equation}
	R_{lcs} = \max_{j=1, ..., u} \left( \frac{LCS(r_j, c)}{m_j} \right)
\end{equation}
and
\begin{equation}
	P_{lcs} = \max_{j=1, ..., u} \left( \frac{LCS(r_j, c)}{n} \right)
\end{equation}
With such results, the LCS-based F-measure - ROUGE-L - can be attained through
\begin{equation}
	F_{lcs} = \frac{(1+\beta^2) R_{lcs} P_{lcs}}{R_{lcs} + \beta^2 P_{lcs}}
\end{equation}
where $\beta = P_{lcs} / R_{lcs}$ when $\partial F_{lcs} / \partial R_{lcs} = \partial F_{lcs} / \partial P_{lcs}$.}

\subsubsection{CIDEr}

{The Consensus-based Image Description Evaluation, (CIDEr)~\cite{vedantam2015cider} is another metric to evaluate the similarity of machine translations in comparison to human-annotated ones. The authors outline three principles with which automatic translations should be evaluated regarding consensus, incorporating them in the calculations shown later. They are:
	\begin{itemize}
		\item A measure of consensus should account for the frequency in which candidate $n$-grams appear in their reference counterparts in the same image;
		\item A measure of consensus should penalize candidate $n$-grams not present in references for the same image; and
		\item A measure of consensus should weight $n$-grams inversely to how frequent they are across all images.
	\end{itemize}
In order to account for these necessities, the authors of \cite{vedantam2015cider} developed the following metric. Let $I_i$ be the $i$-th image in the dataset $I$, with a candidate (machine-generated) sentence represented as $c_i$, and $S_i = \{ s_{i1}, s_{i2}, ..., s_{im} \}$ be the set of all reference sentences. For a particular image, $h_k(s_{ij})$ denotes the the number of times an $n$-gram $\omega_k$ occurs in reference $s_{ij}$ - and for candidates, $h_k(c_i)$. Therefore, the Term Frequency Inverse Document Frequency (TF-IDF)~\cite{robertson2004understanding}, $g_k(s_{ij})$, is calculated:
\begin{equation}
	g_k(s_{ij}) = \frac{h_k(s_{ij})}{\sum_{\omega_l \in \Omega} h_l(s_{ij})} \log \left( \frac{|I|}{\sum_{I_p \in I} \min(1, \sum_q h_k(s_{pq}))} \right)
\end{equation}
where $\Omega$ is the vocabulary of all $n$-grams. With the TF-IDF calculated for every $n$-gram, the vectors $\boldsymbol{g^n}(c_i)$ and $\boldsymbol{g^n}(s_{ij})$, whose elements are $g_k(c_i)$ and $g_k(s_{ij})$ respectively, are used to calculate the CIDEr score for $n$-grams of size $n$:
\begin{equation}
	\text{\textbf{CIDEr}}_n(c_i, S_i) = \frac{1}{m} \sum_j \frac{
		\boldsymbol{g^n}(c_i) \cdot \boldsymbol{g^n}(s_{ij})
	}{
		\lVert \boldsymbol{g^n}(c_i)  \rVert \cdot \boldsymbol{g^n}(s_{ij}) \rVert
	}
\end{equation}
with $\lVert \cdot \rVert$ denoting the magnitude. Finally, the global CIDEr score is calculated:
\begin{equation}
	\text{\textbf{CIDEr}}(c_i, S_i) = \sum_{n=1}^N w_n \text{\textbf{CIDEr}}_n(c_i, S_i)
\end{equation} The authors also propose that uniform weights (i.e. $w = \frac{1}{N}$) are better.}

\subsubsection{Hit@K}

{ One of the known approaches to handle VQA is using Knowledge Graphs, specially with questions where the answer cannot be found in the image but depends on external knowledge~\cite{chen2021zero}. Knowledge Graphs consist of a set of triples $(h, t, r)$ where $h$ and $t$ are entities and $r$ represents the relationship between them.

Hit@K is a metric that gives information on how well a solution is performing~\cite{ali2020metricHitAtK}. It signalizes the rate of correct test triples that have been ranked in the top k triples, and it is given by:

\begin{equation}
Hit@K = \frac{\vert \{  t \in \mathcal{K}_{test} | rank(t) \leq K \}\vert}{\vert \mathcal{K}_{test}\vert} 
\end{equation}
 where $\mathcal{K}_{test}$ is a set of test triples, $h, t \in \mathcal{E}, r \in  \mathcal{R}, (h,t,r) \in \mathcal{K} $ and $ \mathcal{K} \subseteq \mathcal{E} \times \mathcal{E} \times \mathcal{R}$
}
\subsection{Related datasets and metrics}
\label{ss.relatedmethodology}

{Although the majority of works consider only a specific subset of datasets and metrics, usually the most common in VQA literature, many others exist to test certain aspects necessary for a well-rounded VQA solution. In this section, we discuss such datasets and metrics that do not appear as often as the above-mentioned ones, focusing on two observable trends in the works reviewed: bias reduction and external knowledge incorporation.}

\subsubsection{CLEVR-X: A Visual Reasoning Dataset for Natural Language Explanations}

{The article \cite{Salewski2022} proposes an increment to the CLEVR dataset, the CLEVR-X. This increment consists of the insertion of a more described context to all question-answer pairs in the dataset. In order to maintain a reasonable dataset size, filters are applied to keep the descriptions according to their relevance.

The descriptions are generated sequentially: for each image, all the question-answer pairs are mapped; with the local contextualization of each image, filters are applied in order to limit the generated description, keeping only descriptions related in the context of the question-answer pairs; finally, the context text is generated keeping the description consistent with the question-answer pairs. In order to reduce the sizes of the generated texts, descriptions are aggregated numerically.

This study points to good results in easy questions or involving categories. One of the main problems, and consequently a possibility of further studies, is precisely the generation of descriptions related to counting.}

\subsubsection{External Knowledge-related metrics}

{Some papers, as will be elaborated in section \ref{s.discussion}, increment external knowledge in the VQA task in order to make it more robust and attentive to information about the world rather than only what is accessible inside the image. Usually, to model relationships between entities, such as how an object correlates to an actor, a graph structure is developed and refined through learning parameters. Because of this graph dependency of the main VQA model to correctly answer questions, some metrics commonly applied in recommender systems (another graph-heavy task) can be used to evaluate how adequate the retrieval of a certain information is to the final answer.

The most common metrics seen in the reviewed papers that emerge from this sub-task are:
\begin{itemize}
	\item \textit{Precision@K}: When \textit{K} is the number of top items retrieved from outside knowledge to answer a given question, the \textit{Precision@K} measures the rate of relevant ones in the total retrieved ones;
	\item \textit{Recall@K}: Similar to the above, but measures the rate of relevant items retrieved in the total possible relevant items available;
\end{itemize}}

\subsubsection{Bias Reduction-related metrics}

{Another point of discussion in section \ref{s.discussion}, techniques to reduce bias learning in models are now necessary in VQA applications and studies. Although most authors simply apply a module to refrain models from learning spurious correlation in the data~\cite{yang2020learning, wen2021debiased}, or develop new loss functions to achieve the same result~\cite{yang2020learning, Liang2021}, the work \cite{kervadec2021roses} develops a novel metric to measure the accuracy in out-of-distribution data samples, named \textit{Acc-tail}. In said work, this metric measures the prediction accuracy of samples whose class amounts lower than the average of all samples' classes. Its calculation is the same as accuracy, but measured specifically for an OOD setting.

At the intersection of bias reduction and visual grounding, \cite{Han2021} develops new metrics to assess a model's capacity to properly ground their predictions, using the previously proposed Correctly Predicted but Improperly Grounded (CPIG)~\cite{cpigpaper} metric, which is calculated by
\begin{equation}
	\%CPIG = \frac{N_{\text{correct answer, improper grounding}}}{N_{\text{correct answer}}} \times 100\%
\end{equation} where the numerator of the fraction is the number of instances in which the regions used by the model to correctly predict the answer are not in the top-3 most relevant regions. With this, \cite{Han2021} defines three other metrics: Correct Grounding for Right Prediction (CGR) - the complementary opposite of CPIG -, Correct Grounding but Wrong Prediction (CGW), and Correct Grounding Difference (CGD) - the difference of the previous two. These metrics serve to precisely observe how much models are answering correctly despite focusing on visual regions with little to no information, implying a language bias.}

\section{Results}
\label{s.results}

{
In this section, we compare the results obtained in the works reviewed on a quantitative basis. This comparison provides a way to evaluate the most successful contributions in solving the task of VQA and to highlight possible areas of improvement. Since the field of VQA contains many datasets, each with their own unique characteristics, the results are compared considering the dataset evaluated in each work. Table \ref{t.results-acc} ranks them according to their accuracy results.
}

{
For the sake of name simplification, the base model used in \cite{teney2020unshuffling} - deeply explained and detailed in Teney et al.~\cite{teneynet} - will be referred to as TeneyNet. Similarly, for the work of \cite{farinhas2021multimodal}, which implements multiple continuous attention heads on top of a base model, the method will be referred to as the modality of attention evaluated.
}

\begin{longtable}{lrp{2cm}p{2.5cm}}
	
	\toprule
	 &         &         &     \\
	Method     & Dataset          & Base model    &    Accuracy        \\

	\midrule 
	
	GGE-DQ-tog~\cite{Han2021}        & VQA v2     &  UpDn      & $59.11$\footnotemark[1]            \\ 
	CCB + VQ-CSS~\cite{yang2020learning}     &   VQA v2    &     UpDn      &  $59.17$\footnotemark[1]            \\
	GGE-DQ-iter~\cite{Han2021}        & VQA v2     &  UpDn     & $59.30$\footnotemark[1]            \\ 
	CCB~\cite{yang2020learning}     &   VQA v2    &     UpDn      &  $60.73$\footnotemark[1]            \\
	LPF~\cite{Liang2021}         & VQA v2    & UpDn      & $62.63$\footnotemark[1]          \\
	AttReg~\cite{Liu2022}                &  VQA v2       &    LMH    & $62.74$\footnotemark[1] | $63.43$   \\
	TrainEnv (Ensemble)~\cite{teney2020unshuffling}     & VQA v2     &  TeneyNet     &    $63.47$\footnotemark[1]      \\
	D-VQA~\cite{wen2021debiased}     &  VQA v2   &   UpDn    & $64.96$\footnotemark[1]               \\
	AttReg~\cite{Liu2022}                &   VQA v2      &    UpDn    & $64.13$\footnote{\label{vqav2-val}VQA v2 validation data} | $65.25$      \\
	MIRTT~\cite{wang2021mirtt}    &  VQA v2    &  ViLBERT     & $68.0$                 \\
	PointNet~\cite{spencer2021skill}   & VQA v2   & PointNet      & $69.93 \pm 0.155$      \\ 
	OracleTR+LXMERT~\cite{kervadec2021transferable} & VQA v2 & LXMERT & $70.2$ \\
	MIRTT (Ensemble)~\cite{wang2021mirtt}    &  VQA v2    & ViLBERT     & $70.3$                 \\
	UnimodalAtt~\cite{farinhas2021multimodal}    & VQA v2    &  MCAN     & $70.57 \pm 0.16$              \\
	MultimodalAtt~\cite{farinhas2021multimodal}    & VQA v2    &  MCAN   & $70.60 \pm 0.185$              \\
	DiscreteAtt~\cite{farinhas2021multimodal}    & VQA v2    &  MCAN     & $70.76 \pm 0.175$              \\
	CFR-VQA~\cite{nguyen2022coarse}         &  VQA v2     &   CFR         & $69.7$\footnotemark[1] | $72.5$ \\ 
	VLMo~\cite{https://doi.org/10.48550/arxiv.2111.02358}        & VQA v2    &  VLMo-Base     & $76.76 \pm 0.125$            \\
	VLMo~\cite{https://doi.org/10.48550/arxiv.2111.02358}        & VQA v2    &  VLMo-L     & $79.96 \pm 0.02$             \\
	VLMo~\cite{https://doi.org/10.48550/arxiv.2111.02358}        & VQA v2    &  VLMo-L++     & $82.83 \pm 0.05$             \\

	\midrule 
	
	PointNet~\cite{spencer2021skill}   & VQA-CP   & PointNet      & $41.71$       \\
	AttReg~\cite{Liu2022}                &  VQA-CP      &  UpDn       & $47.66$           \\
	AttReg~\cite{Liu2022}                &  VQA-CP      &  LMH       & $62.25$           \\

	\midrule 
	
	AttReg~\cite{Liu2022}   \hspace{4cm}             &  VQA-CP2      &  UpDn      & $46.75$           \\
	TrainEnv~\cite{teney2020unshuffling}     &    VQA-CP2   &  TeneyNet    & $48.06$         \\
	LPF~\cite{Liang2021}         & VQA-CP2    & BAN      & $50.76$          \\ 
	LPF~\cite{Liang2021}         & VQA-CP2    & S-MRL      & $53.38$          \\
	D-VQA~\cite{wen2021debiased}     & VQA-CP2    &   SAN    & $54.39$               \\ 
	LPF~\cite{Liang2021}         & VQA-CP2    & UpDn      & $55.34$          \\ 
	GGE-DQ-iter~\cite{Han2021}        & VQA-CP2    &  UpDn     & $57.12$            \\
	GGE-DQ-tog~\cite{Han2021}         & VQA-CP2     &  UpDn     & $57.32$            \\ 
	CCB~\cite{yang2020learning}     &  VQA-CP2      &   UpDn   &  $57.99$            \\
	CCB + VQ-CSS~\cite{yang2020learning}     &   VQA-CP2    &     UpDn      &  $59.12$            \\
	AttReg~\cite{Liu2022}                &  VQA-CP2      &  LMH       & $60.00$           \\
	D-VQA~\cite{wen2021debiased}     &  VQA-CP2   &   UpDn    & $61.91$    \\
	D-VQA~\cite{wen2021debiased}     &  VQA-CP2   &   LXMERT    & $69.75$  \\

	\midrule 
	
	MIRTT~\cite{wang2021mirtt}    &  GQA    &  SAN     & $52.4$                 \\
	MIRTT (Ensemble)~\cite{wang2021mirtt}    &  GQA    & SAN     & $55.9$                \\
	GraphHopper~\cite{rajar2021graphhopper}      &   GQA      &   GAT     & $56.69$            \\
	OracleTR+LXMERT~\cite{kervadec2021transferable} & GQA &  LXMERT & $57.8$ \\
	CFR-VQA~\cite{nguyen2022coarse}         &  GQA     &     CFR       & $73.6$\footnote{\label{gqa-val}GQA validation data} | $72.1$ \\
	GraphHopper-pr~\cite{rajar2021graphhopper}      &   GQA      &   GAT     & $81.41$            \\

	\midrule 
	
	LaTr-S+Rosetta~\cite{LaTr} & TextVQA & ViT + T5 & $41.84$ \\
	LaTr-B+Rosetta~\cite{LaTr} & TextVQA & ViT + T5 & $44.06$ \\
	SMA+Pretraining~\cite{gao2021structured} & TextVQA & Faster-RCNN & $45.51$ \\
	LaTr-B+Rosetta~\footnote{Pre-trained on IDL data}~\cite{LaTr} & TextVQA & ViT + T5 & $48.38$ \\

	\midrule 
	
	MedVQA~\cite{zhan2020medvqa}            & VQA-RAD            &    BAN + LSTM        & $60.0$              \\
	CPRD + BAN~\cite{liu2021constrative}     & VQA-RAD       &  BAN        &  $67.8$          \\
	CPRD + BAN + CR~\cite{liu2021constrative}     & VQA-RAD       &  BAN + CR        &  $72.7$          \\
	MedVQA~\cite{zhan2020medvqa}            & VQA-RAD            &    BAN + LSTM        & $79.3$\footnote{Evaluation on closed-end answers}              \\

	\bottomrule
	
	\caption{Accuracy of some methodologies reviewed in this survey}
	\label{t.results-acc}
	
\end{longtable}

{ 
Considering Table \ref{t.results-acc}, some explanation is due in order to shed the proper light on certain works. The paper \cite{spencer2021skill} proposes a novel way to evaluate VQA, focusing on skill-concepts (such as animal, color, etc) instead of an overall quantification. The work of \cite{farinhas2021multimodal} implements continuous attention mechanisms in opposition to the common, discrete ones. According to the authors, although the results obtained in DiscreteAtt are better than the continuous methods, the computational efficiency is greater in the latter. Finally, \cite{wang2021mirtt} uses many different base models when performing an ablation study on the applicability of their proposition; we reported only the highest accuracy present in such comparisons. We refer the reader to all the mentioned papers for more details.
}

{
Out of the 25 papers reviewed in this survey, 11 works (44\%) use VQA v2, 2 (8\%) use VQA-CP, 6 (24\%) use VQA-CP2, 3 (12\%) use GQA and 2 (8\%) use VQA-RAD - with some works covering more than one dataset. Besides these, there are 20 datasets used only once in the entirety of the survey, such as VQA-HAT, F-VQA, CLEVR, etc. Five of the covered papers introduce new datasets. Regarding metrics, 14 (56\%) use accuracy as their evaluation metric. 2 of them (8\%) use BLEU and the same 2 use CIDEr. 
}


\section{Discussion}
\label{s.discussion}

{In this section, we provide a discussion concerning the state-of-the-art in VQA, common shortcomings found in the papers reviewed, highlight tendencies observed in recent works, and elaborate on future directions for this research field.}

\subsection{State of the Art (SOTA)}
\label{ss.sota}

{Finding the best techniques and metrics to demonstrate the state-of-the-art is not something easy. To facilitate this, Table \ref{t.results-acc}, which is a compilation of the datasets, metrics and results of the articles used in this survey, was created with extensive research of the topic. According to the research made, it can be observed that two of the most used datasets for Visual Question Answering (VQA) were VQA v2~\cite{vqav2} and VQA-CP2~\cite{vqacpv2}. 
	
	As of the publication of this survey, the model that achieved the best results using the VQA v2 dataset in its test-std version with an open source code was \textbf{BeiT-3} \cite{beit-sota}, using Accuracy as the metric. With its test-dev version, the PaLI~\cite{PaLI} model is the one with highest accuracy. However, the article with the proposed model does not meet the requirements in this survey from section \ref{ss.scope_of_this_work}. It achieved 84.30 of accuracy while BeiT-3 stayed in second place with 84.19. 
	
	Also, when using the VQA-CP2 dataset, the model with the best results was D-VQA \cite{wen2021debiased} when applied to different backbones of VQA models. Using this method, the best result with this dataset and model was when using the LXMERT backbone, followed by the UpDn backbone, both also using Accuracy as the metric. }

\subsection{Tendencies of Recent Works}
\label{ss.tendencies}
In the articles present in this work, some notable tendencies appear in the form of common problems to solve or common approaches to render VQA models more general. In this section, we highlight the most outstanding ones and provide an understanding for them. 

\begin{itemize}
	\item Seven works reviewed focus on the problem of bias learning from spurious correlations within datasets~\cite{yang2020learning, kervadec2021roses, teney2020unshuffling, kervadec2021transferable, wen2021debiased, Han2021, jaunet2021visqa}. VQA datasets are notorious for the biases they contain, leading models to answer, e.g., that "a banana" is "yellow" without actually paying attention to the visual information available. Therefore, to mitigate this problem and make models confidently rely on visual queues, some authors focus on redistributing data within datasets to increase the need for models to learn how to articulate questions and answers~\cite{kervadec2021roses, teney2020unshuffling}. Others, however, implement new modules and branches for models in order to focus on the most important data present in image-question pairs~\cite{yang2020learning, wen2021debiased, Han2021, jaunet2021visqa}.
	\item Another form of increasing VQA models' generalization capabilities is the introduction of external knowledge in the models' inference process. With this addition, the learning process can be guided by information provided from experts and makes the model more robust.
	\item Also relating to the aforementioned external knowledge injection approach, graph learning can be used to make VQA models associate information more precisely. Through the use of knowledge graphs to create relationships between objects and subjects within an image, models can answer more confidently about the context of some information present in the image, rendering them more robust \cite{chen2021zero, rajar2021graphhopper, ali2020metricHitAtK}. 
\end{itemize}

\subsection{Common Problems}
\label{ss.common_problems}

To give an evaluation about the articles and analysis to future works, this section highlights some observed problems in the works reviewed:
\begin{itemize}
	
\item{Firstly, the VQA task is a wild area with multiple niches of knowledge, such as medical, educational, visual-impairment assistance, and others. Because of that, the datasets sometimes will not contemplate all works because of the specificity needed to build a model.}
\item{Following the discussion above, it is common for authors to develop an entirely new dataset by themselves, causing a non-standardization in the field.}
\item{In addition, VQA models tend to leverage huge architectures with many training parameters, consequently requiring lots of computational resources in calculation-intensive training procedures, rendering a widespread scientific focus impossible for the subject.}
\item{Finally, in the last few years, most articles focus on the tendencies and stopped focusing in other areas, creating a 'time-stop' and making it harder to evaluate those other areas.}
\end{itemize}

\subsection{Common Weaknesses}
\label{ss.common_weaknesses}

Following subsection \ref{ss.common_problemns}, the weaknesses analyzed in the articles will provide an alternative to continue works in the future:
\begin{itemize}
\item{Due to the tendency of articles, if someone looks for a different subject within the area, they are unlikely to find many recent works.}
\item{Although the name of this task is Visual Question Answering, some articles do not provide visual context to facilitate the reading.}
\end{itemize}

\section{Conclusion}
\label{s.conclusion}

{
Although VQA models have had a huge progress when compared to previous years, there is still space for improvement.
Nevertheless, VQA is an area of great opportunity, and can be used in medicine, robotics, security, welfare, etc.This is possible by the use of Deep Learning, which in general shows a lot of advances for recognizing and evaluating images and texts together. In addition, this survey introduces basic concepts about VQA, with an evaluation of the most recent papers about the subject, and an overview of the field.
This survey includes 29 papers, selected based on the following criteria: their content, their open access and open source situation, and them being fairly recent (published in the last two years), creating an ultra-recent survey with the state-of-the-art.
During the evaluation of the reviewed works, some crucial points were detected:}
	
	\begin{itemize}
		\item The Accuracy was the most used evaluation method.
		\item CNN, RNN and Transformers were the most used machine learning architectures.
	\end{itemize}
	
To further enrich the field of VQA and deep learning applications in medicine, a highlight of some important points for future works is made:
	\begin{itemize}
		\item The creation of a dense and plural common database with images and texts exclusive to VQA.
		\item New works can be created using the papers presented in this work, and become the state-of-the-art.
	\end{itemize}

\section*{Acknowledgments}
The authors are grateful to the Eldorado Research Institute.

\bibliographystyle{unsrt}  
\bibliography{sections/refs}

\end{document}